%% file: main.tex
\documentclass[10pt,twocolumn,letterpaper]{article}
\newcommand{\DATA}{\textsc{CompositionCap}\xspace}
\newcommand{\NAME}{\textsc{FineCaption}\xspace}
 \usepackage[pagenumbers]{cvpr} 

\definecolor{cvprblue}{rgb}{0.21,0.49,0.74}

\def\paperID{11933} 
\def\confName{CVPR}
\def\confYear{2025}

\usepackage{inconsolata}
\usepackage{makecell}
\usepackage{svg}
\usepackage{times}
\usepackage{latexsym}
\usepackage{graphicx}
\usepackage{tabularx}
\usepackage{xspace}
\usepackage{times}
\usepackage{booktabs}
\usepackage{latexsym}
\usepackage{graphicx}
\usepackage{multirow}
\usepackage{multicol}
\usepackage{listings}
\usepackage{amsmath}
\usepackage{makecell}
\usepackage{algorithm}
\usepackage{algpseudocode} 
\usepackage{mathtools}

\usepackage{pifont}
\usepackage{xcolor}
\usepackage{tcolorbox}
\usepackage{colortbl}
\usepackage{dashrule}

\usepackage{comment}

\definecolor{url_color}{RGB}{153, 102,  0}
\definecolor{mygray-bg}{gray}{0.9}
\definecolor{api}{HTML}{ECF4FF}
\definecolor{my_green}{RGB}{51,102,0}
\definecolor{my_red}{RGB}{204, 0, 0}
\usepackage{microtype}

\definecolor{SP}{RGB}{114, 97, 171}
\definecolor{my_green}{RGB}{51,102,0}
\definecolor{my_red}{RGB}{204, 0, 0}
\definecolor{api}{HTML}{FFF9E3}

\newcommand{\cmark}{\textcolor{my_green}{\ding{51}}} 
\newcommand{\xmark}{\textcolor{my_red}{\ding{55}}} 
\usepackage[pagebackref,breaklinks,colorlinks,allcolors=cvprblue]{hyperref}
\title{\NAME: Compositional Image Captioning Focusing on Wherever You Want at Any Granularity}

\author{
     Hang Hua$^{1}$,\quad
     Qing Liu$^{2}$,\quad 
     Lingzhi Zhang$^{2}$,\quad 
     Jing Shi$^{2}$,\quad 
     Soo Ye Kim$^{2}$,\quad 
     Zhifei Zhang$^{2}$, \\
     Yilin Wang$^{2}$, \quad
     Jianming Zhang$^{2}$, \quad
     Zhe Lin$^{2}$, \quad
     Jiebo Luo$^{1}$\\
     $^1$University of Rochester,\quad
     $^2$Adobe Research\\
     {\tt\small \{hhua2,jluo\}@cs.rochester.edu}, {\tt\small \{qingl,lingzzha,jingshi,sooyek,zzhang,yilwang,zlin\}@adobe.com}
}

\begin{document}
\twocolumn[{%
\renewcommand\twocolumn[1][]{#1}%
\maketitle
\vspace{-8mm}
\begin{center}
\centering
\includegraphics[width=0.96\linewidth]{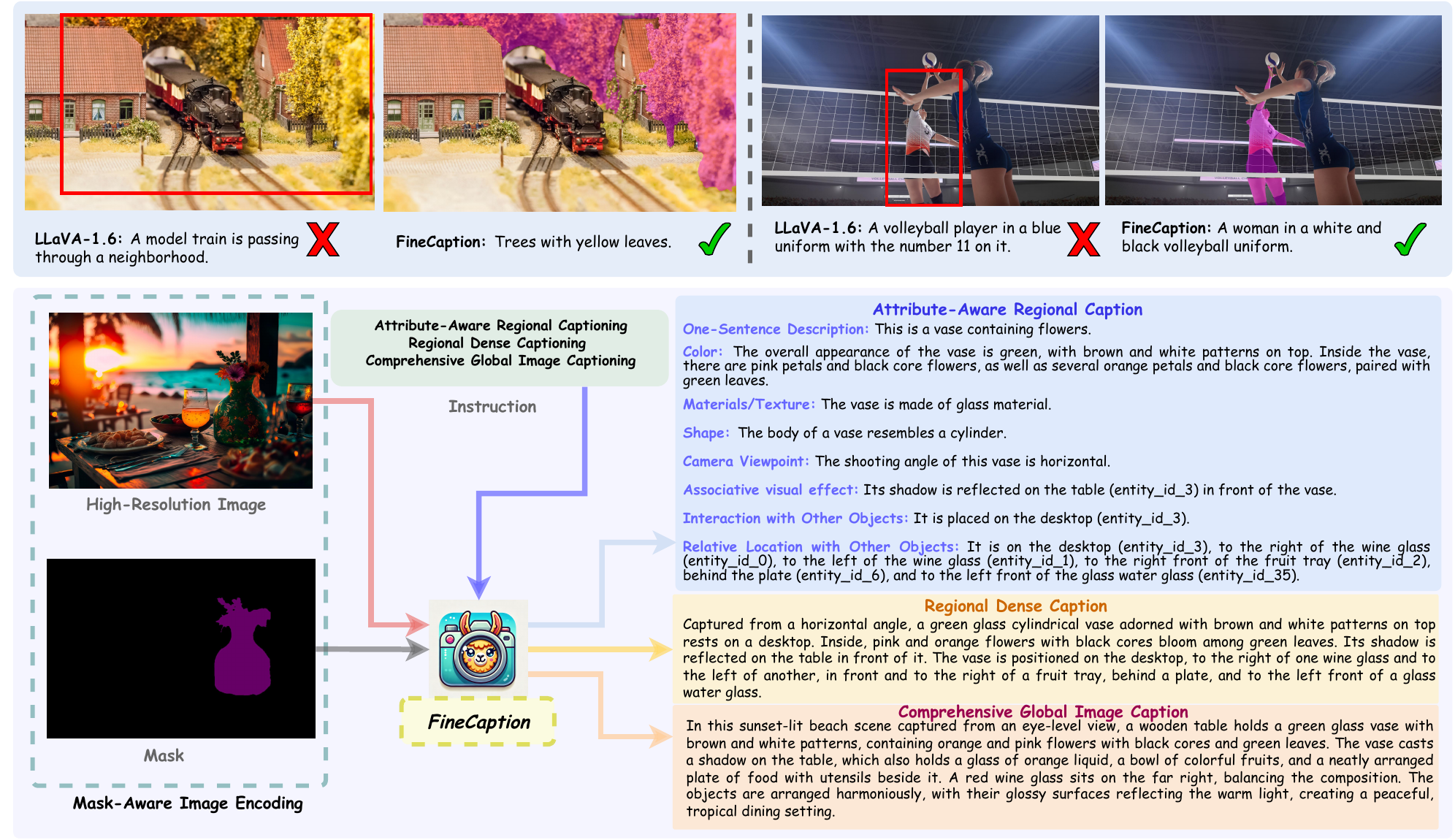}
\vspace{-2mm}
\captionof{figure}{We propose \textbf{\NAME}, a novel Vision-Language model with the improved capabilities of \textbf{Attribute-Aware Regional Captioning}, 
\textbf{Regional Dense Captioning}, and 
\textbf{Comprehensive Global Image Captioning}. \NAME can recognize arbitrary masks as referential inputs and process high-resolution images. Moreover, models trained using the traditional bounding boxes as region reference are inadequate to precisely describe the region of interest.}
\label{fig:teaser}
\vspace{-1mm}
\end{center}%
}]

\maketitle
\input{sec/0_abstract}
\input{sec/1_intro}
\input{sec/2_relatedwork}
\input{sec/3_dataset}
\input{sec/3_method}

\input{sec/4_experiment}

\input{sec/5_analysis}
\input{sec/6_conclusion}

{
    \small
    \bibliographystyle{ieeenat_fullname}
    \bibliography{main}
}
\input{sec/X_suppl}

\end{document}

%% file: sec/0_abstract.tex
\begin{abstract}
The advent of large Vision-Language Models (VLMs) has significantly advanced multimodal tasks, enabling more sophisticated and accurate reasoning across various applications, including image and video captioning, visual question answering, and cross-modal retrieval.
Despite their superior capabilities, VLMs struggle with fine-grained image regional composition information perception. Specifically, they have difficulty accurately aligning the segmentation masks with the corresponding semantics and precisely describing the compositional aspects of the referred regions. However, compositionality -- the ability to understand and generate novel combinations of known visual and textual components -- is critical for facilitating coherent reasoning and understanding across modalities by VLMs. To address this issue, we propose \NAME, a novel VLM that can recognize arbitrary masks as referential inputs and process
high-resolution images for compositional image captioning at different granularity levels. To support this endeavor, we introduce \DATA, a new dataset for multi-grained region compositional image captioning, which introduces the task of compositional attribute-aware regional image captioning. Empirical results demonstrate the effectiveness of our proposed model compared to other state-of-the-art VLMs. Additionally, we analyze the capabilities of current VLMs in recognizing various visual prompts for compositional region image captioning, highlighting areas for improvement in VLM design and training. \url{https://hanghuacs.github.io/FineCaption/}
\end{abstract}

%% file: sec/1_intro.tex
\section{Introduction}
\label{sec:intro}
Pre-trained vision-language models, such as GPT-4o \citep{achiam2023gpt}, LLaVA \citep{liu2024visual}, InternVL \citep{chen2024far}, BLIP \cite{li2023blip}, and VILA \citep{lin2023vila}, have demonstrated impressive capabilities in complex reasoning, and have achieved remarkable results in various vision-language (VL) tasks. Among these tasks, one fundamental challenge is detailed image content perception and captioning, which involves understanding visual inputs and generating descriptive textual outputs. This capability is essential for applications such as assistive technologies, content accessibility, and enhanced human-computer interaction. While extensive research has focused on improving the captioning abilities of VLMs in areas like generic image captioning \citep{zeng2024meacap,you2016image,li2024evcap,wang2022git,ye2023mplug}, dense image captioning \citep{onoe2024docci,johnson2016densecap,wu2025grit}, and referring image captioning \citep{zhao2025controlcap,you2023ferret,sun2024alpha,yuan2024osprey,peng2023kosmos}, less attention has been devoted to region compositional captioning and detailed region-level captioning tasks. These tasks require models to recognize both compositional attribute prompts and region prompts, generating captions specific to the attributes and regions of interest.

Previous works have aimed at enabling region-level understanding in VLMs. Methods such as Kosmos-2 \cite{peng2023kosmos}, Shikra \cite{chen2023shikra}, and GPT4RoI \cite{zhang2023gpt4roi} have attempted to process regions specified by bounding boxes, leveraging visual instruction tuning with object-level spatial features. Other approaches aim to enable VLMs to recognize arbitrary visual prompts and focus on the referred regions by incorporating overlaid images and visual prompts as input \cite{cai2024vip,you2023ferret}. However, these methods have limitations. As illustrated in previous work \cite{sun2024alpha,rasheed2024glamm,zhang2024omg,yuan2024osprey} and the examples in Figure \ref{fig:teaser}, bounding boxes are inadequate for providing precise references to image regions (The IoU between masks and their bounding boxes is 56.11 for our data.). Moreover, freeform overlaid visual prompts are suboptimal for region-level understanding tasks, as they often confuse VLMs; models frequently interpret the visual prompts as integral parts of the image's semantic content. Therefore, utilizing masks as region references is an ideal solution for regional understanding tasks.

In this study, we address these limitations from both the model design and dataset construction perspectives. Specifically, we propose a novel VLM, \NAME, capable of multi-grained region compositional attribute captioning. To better capture detailed compositional information and accurately recognize mask-referred regions, we design a new architecture that integrates a mask-aware low-resolution encoder with multiple high-resolution encoders. Table \ref{tab:model_compare} summarizes the distinct capabilities of our model compared to existing models for regional image understanding. For the mask-aware encoding, we follow the method of Alpha-CLIP \cite{sun2024alpha} by introducing an extra convolution layer to the CLIP image encoder, which incorporates binary masks as the alpha channel for RGB images. Our experiments demonstrate that aggregating multiple high-resolution encoders enhances the model's ability to perceive detailed information in image regions. Therefore, we utilize ConvNeXT \cite{liu2022convnet} and the SAM encoder \cite{kirillov2023segment} as high-resolution encoders, supporting image encoding at the 1024 x 1024 resolution. Furthermore, we introduce a new human-annotated, high-quality dataset, \DATA, to improve models' capabilities in \textbf{Attribute-Aware Regional Captioning (AARC)}, \textbf{Regional Dense Captioning (RDC)}, and \textbf{Comprehensive Global Image Captioning (CGIC)}. Our dataset encompasses diverse scenes and introduces the task of compositional aspect-aware regional image captioning, including 18 different compositional attributes. Detailed explanations of these aspects are provided in the appendix.

Unlike the Referring Expression Generation (REG) task -- which involves automatically generating a referring expression that uniquely identifies a particular object in an image by distinguishing it from other objects in the same scene -- \DATA emphasizes the generation of multi-grained compositional expressions for regions of interest. In this task, models are required to describe masked regions with detailed and comprehensive compositional information. Our objective is not to differentiate objects from one another but rather to provide rich, attribute-aware descriptions of the specified regions \cite{yu2016modeling,kazemzadeh2014referitgame}. The goal is to create captions that not only mention objects or regions but also elaborate on their compositional aspects and how they interact or relate to the broader scene. 

\input{tables/model_comparison}
 Empirical results demonstrate the superior performance of \NAME on region compositional image captioning tasks compared to other strong VLMs, especially GPT-4 \cite{achiam2023gpt}, and LLaMA-3.2 \cite{dubey2024llama} which is recognized for its strong capabilities across a wide range of tasks.
 
In summary, our contributions are threefold:
\begin{itemize}
    \item We propose \NAME, a novel vision-language model with enhanced capabilities for mask-referring multi-grained image compositional captioning. \NAME is equipped with a mask-aware image encoder for mask-referring and a high-resolution encoder for fine-grained perception of compositional information. Empirical results demonstrate the superior performance of our model on mask-referring image compositional captioning tasks compared to other strong VLMs.
    \item We propose \DATA, a novel, human-annotated, high-quality benchmark for multi-grained mask-referring image compositional captioning. \DATA encompasses 18 different compositional aspects and offers three levels of captioning granularity: Attribute-Aware Regional
Captioning, Regional Dense Captioning, and Comprehensive Global Image Captioning.
    \item We analyze the capabilities of VLMs in generating compositional aspect-aware captions for detailed region descriptions and in handling input images with region-referring tasks. This analysis highlights key areas where models can improve in capturing compositional attributes -- such as color, body gesture, material, and texture -- and in accurately distinguishing specific regions. These insights offer directions for future development in both caption generation and recognition tasks.
\end{itemize}

%% file: tables/model_comparison.tex
\begin{table*}[t]
\centering
\resizebox{0.95\linewidth}{!}{%
\begin{tabular}{l|ccccc}
\toprule
\textbf{Model} & \textbf{Mask Referencing} & \textbf{High Resolution} & \textbf{Region Attribute Captioning} & \textbf{Region Dense Captioning} \\
\midrule
\midrule
ViP-LLaVA \cite{cai2024vip}&\xmark&\xmark&\xmark&\xmark \\
Ferret \cite{you2023ferret}&\xmark&\xmark&\xmark&\xmark \\
Ferret-v2 \cite{zhang2024ferret}&\xmark&\cmark&\xmark&\xmark \\
GPT4RoI \cite{zhang2023gpt4roi}&\xmark&\xmark&\xmark&\xmark \\
VCoder \cite{jain2024vcoder}&\cmark&\xmark&\xmark&\xmark \\
Osprey \cite{yuan2024osprey}&\cmark&\xmark&\xmark&\cmark\\
Alpha-CLIP \cite{sun2024alpha}&\cmark&\xmark&\xmark&\xmark \\
GLaMM \cite{rasheed2024glamm}&\cmark&\xmark&\cmark&\xmark \\
RegionGPT \cite{guo2024regiongpt}&\cmark&\xmark&\xmark&\xmark \\
OMG-LLaVA \cite{zhang2024omg}& \cmark&\xmark&\xmark&\xmark \\
\midrule
\rowcolor[HTML]{ECF4FF}
\textbf{\NAME (ours)}& \cmark  & \cmark & \cmark & \cmark\\ 
\bottomrule
\end{tabular}%
}
\caption{Comparison of the capabilities of \NAME and other related VLMs: ``Mask Referring'' indicates whether the model's encoder can accept a mask input as a reference, ``High-Resolution'' specifies whether the model supports high-resolution image encoding. }
\label{tab:model_compare}
\vspace{-4mm}
\end{table*}

%% file: sec/2_relatedwork.tex
\section{Related Work}
\label{sec:rwork}

\subsection{Pre-trained Vision-Language Models}
Vision-language models~\cite{radford2021learning,liu2024visual,hua2024v2xum,xu2023mplug,tang2024avicuna,chen2024far,li2022blip,tong2024cambrian,cheng2024spatialrgpt} strive for multimodal intelligence by jointly processing visual and linguistic information. Inspired by the remarkable success of recent large language models (LLMs)~\cite{touvron2023llama,chiang2023vicuna,hua2023improving}, researchers are now exploring large VLMs that combine pre-trained visual encoders and language decoders to tackle complex multimodal tasks. Flamingo~\cite{alayrac2022flamingo} and BLIP-2~\cite{li2023blip} are two of the early works that explore the integration of LLMs into vision-language pre-training. These models are trained as VL foundation models. Beginning with LLaVA~\cite{liu2024visual}, researchers have used LLM-synthesized instruction-following chat data in VQA format for instruction tuning, achieving significantly improved results~\cite{hua2024finematch,hua2024mmcomposition,yu2023mm, tang2024vidcomposition}. Subsequent studies have expanded to explore the broader capabilities~\cite{hu2023promptcap, hua2024mmcomposition, lin2023videoxum, yu2024promptfix} of multimodal LLMs. However, these efforts have placed less emphasis on enhancing models' ability to describe image contents focusing on specific regions and attributes.
\subsection{Vision-Language Models for Region-level Image Understanding}
Recent works like Kosmos-2 \cite{peng2023kosmos}, Shikra \cite{chen2023shikra}, GPT4RoI \cite{zhang2023gpt4roi}, Ferret \cite{you2023ferret}, and Sphinx \cite{lin2023sphinx} aim to enable region-specific interactions in vision-language models (VLMs). A straightforward approach is to provide bounding box coordinates to the models, as employed by Kosmos-2, Shikra, RegionGPT \cite{guo2024regiongpt}, and Sphinx. Methods such as ViP-LLaVA \cite{cai2024vip} and Ferret explore recognizing arbitrary freeform visual prompts by overlaying visual prompts onto images. However, this input format can confuse VLMs, as the models often interpret the visual prompts as part of the image's semantic content. Some models, including Alpha-CLIP \cite{sun2024alpha}, OMG-LLaVA \cite{zhang2024omg}, GLaMM \cite{rasheed2024glamm}, Osprey \cite{yuan2024osprey}, and VCoder \cite{jain2024vcoder}, use masks for region referencing, offering more fine-grained control. However, they process images at resolutions ranging from 224$\times$224 to 448$\times$448, limiting their ability to capture fine-grained compositional details. High-resolution image encoding enhances a model's capacity for detailed perception, which is crucial for understanding and describing complex visual information accurately.


%% file: sec/3_dataset.tex
\section{The \DATA Dataset}
\label{sec:dataset}
To enhance the capabilities of vision-language models (VLMs) in Attribute-Aware Regional Captioning, Regional Dense Captioning, and Global Dense Captioning, we constructed a new human-annotated, high-quality dataset named \DATA. We began by collecting high-quality images from various stock image sources such as Adobe Stock Images and iStock. We then obtained entity masks using the SAM model \cite{kirillov2023segment}. Human annotators were tasked with describing the attributes of these entities to generate corresponding region compositional attribute descriptions. This process resulted in 14,590 entities across 5,392 images and a total of 186,490 attribute descriptions. To enhance models' capabilities in comprehensive global image captioning, we also construct instruction-tuning data based on DOCCI \cite{onoe2024docci}, a dataset containing long, human-annotated English descriptions for 15,000 images.

\begin{figure}[htbp]
    \centering
    \vspace{-2mm}
    \includegraphics[width=0.95\linewidth]{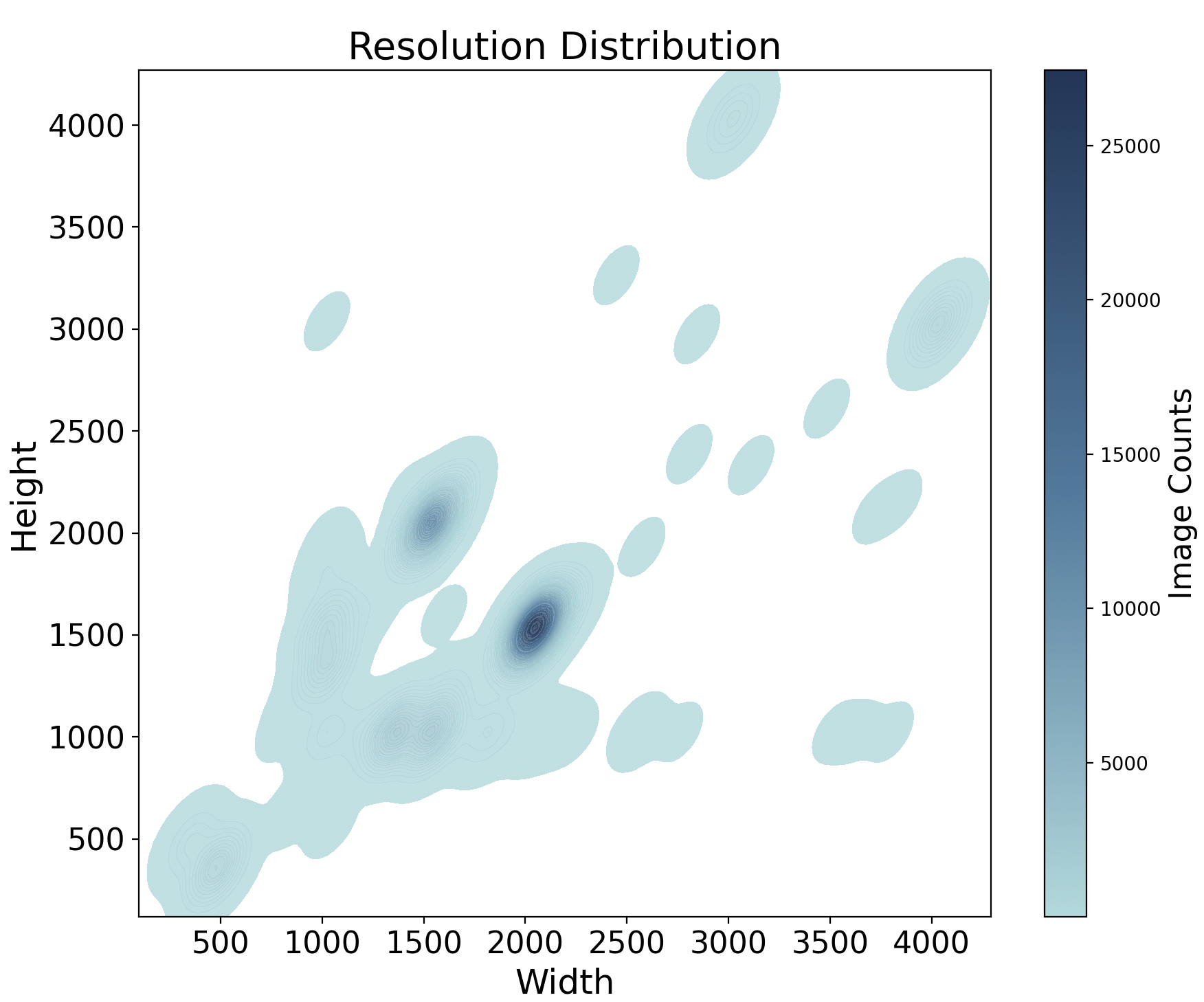}  
    \vspace{-2mm}
    \caption{The image resolution distribution across data points in \DATA reflects the high overall quality of images.}
    \label{fig:resolution}
    \vspace{-4mm}
\end{figure}

\begin{figure}[htbp]
    \centering
    \vspace{-2mm}
    \includegraphics[width=0.95\linewidth]{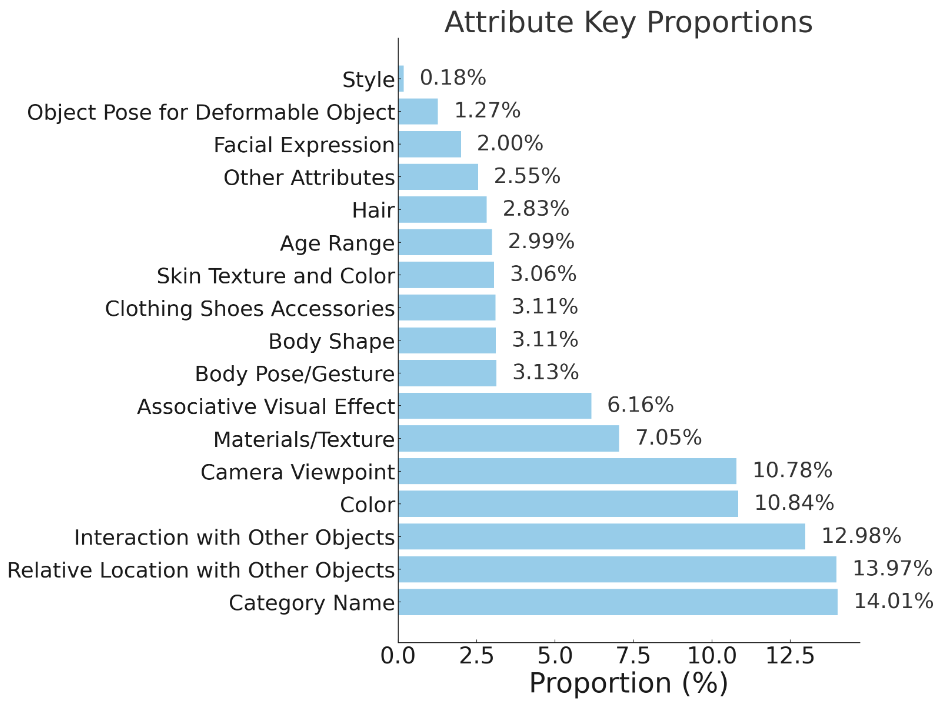}  
    \vspace{-2mm}
    \caption{Distribution of attributes in \DATA.}
    \label{fig:proportion}
\vspace{-4mm}
\end{figure}

In addition, we built out-of-domain test set for \DATA sourced from the Open Images dataset \cite{kuznetsova2020open}. We selected 1,000 images featuring diverse and complex scenes for annotation. The test set contains 7,215 masked entities with 19,326 attribute-specific region captions. The attributes in \DATA include:
1) \textbf{Category Name}, 2) \textbf{Body Shape}, 3) \textbf{Skin Texture and Color}, 4) \textbf{Clothing, Shoes, Accessories}, 5) \textbf{Interaction with Other Objects}, 6) \textbf{Body Pose/Gesture}, 7) \textbf{Other Attributes}, 8) \textbf{Relative Location with Other Objects}, 9) \textbf{Color}, 10) \textbf{Materials/Texture}, 11) \textbf{Camera Viewpoint}, 12) \textbf{Associative Visual Effect}, 13) \textbf{Shape}, 14) \textbf{Facial Expression}, 15) \textbf{Hair}, 16) \textbf{Age Range}, 17) \textbf{Object Pose for Deformable Objects}, 18) \textbf{Style}. Figure \ref{fig:resolution} shows the resolution distribution of \DATA, and Figure \ref{fig:proportion} illustrates the proportion of each attribute in our dataset. More statistical results and examples are included in the appendix, where we also provide a detailed explanation of these attributes.


%% file: sec/3_method.tex
\section{Methodology}
\label{sec:method}
\begin{figure*}[htbp]
    \centering
    \includegraphics[width=\textwidth]{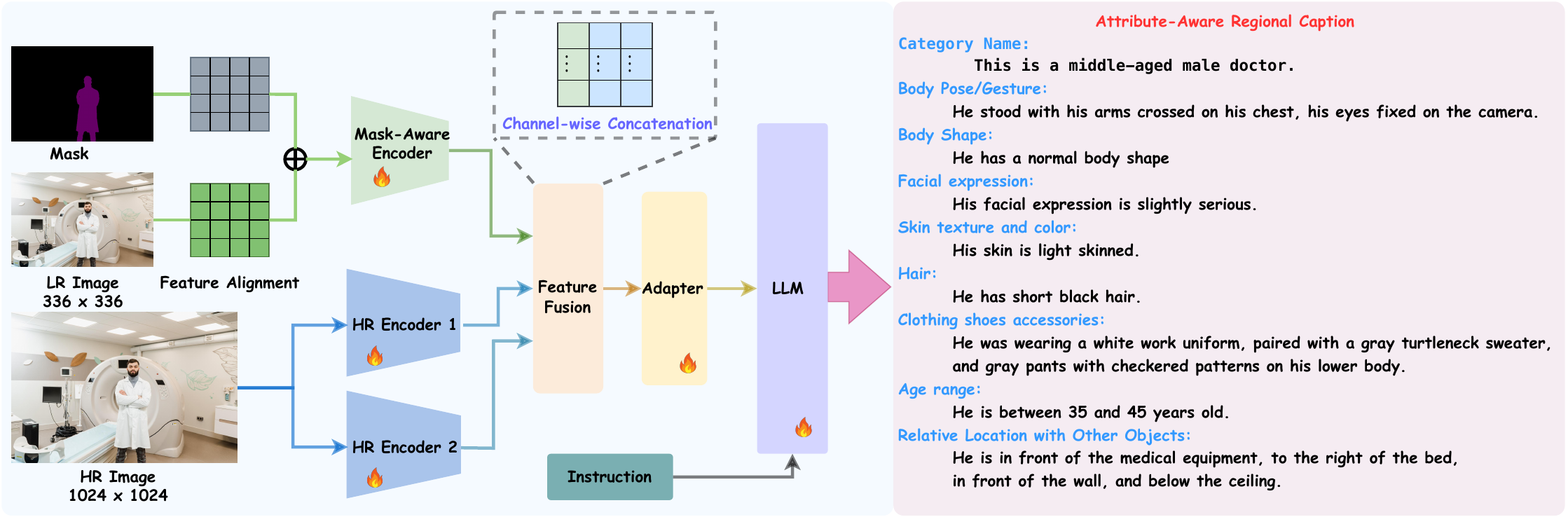}  
    \caption{Overview of \NAME: The model incorporates a mask-aware visual encoder and two high-resolution encoders (ConvNext and SAM), enabling precise recognition of mask references and the perception of detailed compositional and spatial information for images. }
    \label{fig:model_architecture}
    \vspace{-4mm}
\end{figure*}


Our proposed model integrates mask-aware and high-resolution features through a multi-resolution image encoding framework, as shown in Figure~\ref{fig:model_architecture}. This section details each component, including the mask-aware encoder, high-resolution encoders, channel-wise feature fusion, and integration with a large language model via an adapter module.

\subsection{Input Representation}
The input of \NAME encoder includes:
\begin{enumerate}
    \item \textbf{Low-resolution image}: \( I_{\text{LR}} \in \mathbb{R}^{H_{\text{LR}} \times W_{\text{LR}} \times 3} \), where \( H_{\text{LR}} = W_{\text{LR}} = 336 \).
    \item \textbf{High-resolution image}: \( I_{\text{HR}} \in \mathbb{R}^{H_{\text{HR}} \times W_{\text{HR}} \times 3} \), where \( H_{\text{HR}} = W_{\text{HR}} = 1024 \).
    \item \textbf{Binary mask}: \( M \in \mathbb{R}^{H_{\text{LR}} \times W_{\text{LR}}\times 1} \), indicating the region of interest within \( I_{\text{LR}} \).
\end{enumerate}

\subsection{Mask-Aware Encoding}
To align the mask-referred region with the input images, we introduce an additional alpha channel to the CLIP encoder's embedding layer \( \text{Conv}_{\alpha} \), following the approach of Alpha-CLIP~\cite{sun2024alpha}. This method enables us to encode the mask independently, preserving the original content of the images.

First, the low-resolution image \( I_{\text{LR}} \) is passed through the encoder’s standard patch embedding layer:
\begin{equation}
    \mathbf{E}_{\text{patch}} = \text{Conv}_{\text{RGB}}(I_{\text{LR}}),
\end{equation}
where \( \text{Conv}_{\text{patch}}(\cdot) \) maps \( I_{\text{LR}} \) into patch embeddings \( \mathbf{E}_{\text{patch}} \in \mathbb{R}^{C \times H' \times W'} \) (\( C \) is the number of output channels, \( H' \) and \( W' \) are the spatial dimensions of the patch embeddings).

Simultaneously, the mask \( M \) is processed by the additional convolutional layer:
\begin{equation}
    \mathbf{E}_{\text{mask}} = \text{Conv}_{\alpha}(M),
\end{equation}
where \( \text{Conv}_{\alpha}(\cdot) \) is a convolutional layer with the same parameters as \( \text{Conv}_{\text{RGB}} \) except for the input channels, which is set to 1 for the mask. This layer outputs mask embeddings \( \mathbf{E}_{\text{mask}} \in \mathbb{R}^{C \times H' \times W'} \).

The patch embeddings and mask embeddings are then combined and flattened to form the input sequence:
\begin{equation}
    \mathbf{E}_{\text{seq}} =\text{Flatten}(\mathbf{E}_{\text{patch}} + \mathbf{E}_{\text{mask}})^\top \in \mathbb{R}^{N \times C},
\end{equation}
where \( N = H' \times W' \) is the number of patches.

We then add class embeddings \( \mathbf{E}_{\text{class}} \) and positional embeddings \( \mathbf{E}_{\text{pos}} \):
\begin{equation}
    \mathbf{E} = [\mathbf{E}_{\text{class}}; \mathbf{E}_{\text{seq}}] + \mathbf{E}_{\text{pos}},
\end{equation}
resulting in input embeddings for the encoder.

The mask-aware feature map \( \mathbf{F}_{M} \) is obtained by passing \( \mathbf{E} \) through the mask-aware CLIP encoder:
\begin{equation}
    \mathbf{F}_{M} = \text{Encoder}_{M}(\mathbf{E}),
\end{equation}
where \( \mathbf{F}_{M} \in \mathbb{R}^{N+1 \times C_{M}} \) captures the region information of the low-resolution image and the mask.

\subsection{High-Resolution Encoding}
To capture more detailed spatial information for language models, we employ two high-resolution encoders to extract fine-grained features: ConvNext \cite{liu2022convnet} and SAM \cite{kirillov2023segment} encoders.
\begin{align}
    \mathbf{F}_{\text{HR1}} &= \text{Encoder}_{\text{ConvNeXT}}(I_{\text{HR}}), \\
    \mathbf{F}_{\text{HR2}} &= \text{Encoder}_{\text{SAM}}(I_{\text{HR}}),
\end{align}
where \( \mathbf{F}_{\text{HR1}} \in \mathbb{R}^{N' \times C_{\text{HR1}}} \) and \( \mathbf{F}_{\text{HR2}} \in \mathbb{R}^{N' \times C_{\text{HR2}}} \) (\(N' = H'' \times W''\)) are the resulting feature maps from each encoder.

\subsection{Feature Fusion}
The mask-aware and high-resolution features are unified through a channel-wise fusion module. First, the mask-aware feature map \( \mathbf{F}_{M} \) is resized as \( \mathbf{\overline{F}}_{M} \) and then interpolated to match the spatial dimensions of the high-resolution feature maps and finally flattened into a sequence:
\begin{equation}
    \mathbf{F}_{M}' = \text{Flatten}(\text{Interpolate}(\mathbf{\overline{F}}_{M})),
\end{equation}
where \( \mathbf{F}_{M}' \in \mathbb{R}^{N' \times C_{M}} \).

The three feature maps are concatenated along the channel dimension:
\begin{equation}
    \mathbf{F}_{\text{fusion}} = [\mathbf{F}_{M}'; \mathbf{F}_{\text{HR1}}; \mathbf{F}_{\text{HR2}}],
\end{equation}
resulting in \( \mathbf{F}_{\text{fusion}} \in \mathbb{R}^{H'' \times W'' \times C_{\text{fusion}}} \), where \( C_{\text{fusion}} = C_{M} + C_{\text{HR1}} + C_{\text{HR2}} \).

\subsection{Adapter and Language Model Integration}
The fused feature map is then mapped into the LLM's word embedding space by an adapter:
\begin{equation}
    \mathbf{F}_{\text{adapted}} = \text{Adapter}(\mathbf{F}_{\text{fusion}}),
\end{equation}
where \( \mathbf{F}_{\text{adapted}} \in \mathbb{R}^{N' \times D'} \), with \( N' \) and \( D' \) representing the adapted sequence length and embedding dimension for the LLM. These adapted features, along with an instruction \( \mathbf{I} \), guide the LLM’s response generation:
\begin{equation}
    y = \text{LLM}(\mathbf{F}_{\text{adapted}}, \mathbf{I}),
\end{equation}
where \( y \) is the output generated by the LLM, conditioned on the visual features and task-specific instructions.

\subsection{Training Objective}
We then train the model end-to-end using negative log-likelihood loss:
\begin{equation}
    \mathcal{L} = - \sum_{\mathcal{D}} \sum_{i=1}^N  \log p(y_i \mid \mathbf{F}_{\text{adapted}},\mathbf{I}, y_{<i}),
\end{equation}
where \( \mathcal{D} \) represents the training dataset, \( N \) is the total number of tokens or features in each sequence, and \( y_i \) represents the language tokens in the generated output.

\subsection{Training Strategy}
\noindent{\textbf{Stage 1: Pre-training.}}
Similar to LLaVA \cite{liu2024visual}, this stage aims to optimize the projector to align the visual features with the word embeddings of the language model (LLM). Thus, the image encoders and the LLM are kept frozen during pre-training. We use 
LLaVA-Pretrain  \cite{liu2024visual} dataset for training, with the mask in this stage highlighting all regions of the image.

\noindent{\textbf{Stage 2: Image-Mask Alignment Pre-training.}}
The goal of this stage is to align the image and mask features for the mask-aware encoder. We use the data including \DATA, GranD \cite{rasheed2024glamm}, RefCOCO, RefCOCO+ \cite{kazemzadeh2014referitgame}, and RefCOCOg \cite{yu2016modeling} for training. In this stage, only the mask-aware encoder is trainable. 

\input{tables/main_results}

\noindent{\textbf{Stage 3: Region Attribute-Aware Instruction Tuning.}} In the final stage, we use the \DATA training set, allowing all parameters to be trainable. This stage fine-tunes the model to accurately handle compositional referring expressions, enhancing its ability to produce detailed, attribute-aware captions.

%% file: tables/main_results.tex
\begin{table*}[htbp]
\centering

\resizebox{1\linewidth}{!}{%
\begin{tabular}{l|ccc|ccccc}
\toprule
\multirow{2}{*}{\textbf{Model}} & \multicolumn{3}{c|}{\textbf{Region Referral}} & \multicolumn{5}{c}{\textbf{Semantic Evaluation}} \\
\cmidrule{2-9}
& \textbf{Visual Prompt} & \textbf{Resolution} & \textbf{\# Image Token} & \textbf{ROUGE-L}$\uparrow$ & \textbf{BLEU-4}$\uparrow$ & \textbf{METEOR}$\uparrow$ & \textbf{CIDEr}$\uparrow$ & \textbf{BERT Score}$\uparrow$ \\
\midrule
\rowcolor[HTML]{e9edf6}
\multicolumn{9}{c}{\textbf{\textit{Zero-Shot Learning}}} \\
\midrule
Kosmos-2 \cite{peng2023kosmos} & Bbox & 224 & 256 &9.21 &0.14 &1.98 &1.07 & 37.69\\
Alpha-CLIP-13B \cite{sun2024alpha} & Mask & 336 & 576 &13.89 &0.51 &5.94 &2.68 &42.01 \\
Qwen2-VL-7B \cite{Qwen2VL} & Bbox & AnyRes & - &14.12 &0.57 &6.18 &2.74 &42.97 \\
Ferret-13B \cite{you2023ferret} & MContour & 336 & 576 &15.01 &1.06 &5.86 &3.12 &43.82 \\
ViP-LLaVA-13B \cite{cai2024vip} & MContour & 336 & 576 &15.47 &1.48 &5.76 &3.84 &44.29 \\
LLaMA-3.2-11B-Vision-Instruction \cite{dubey2024llama} & Bbox & - & - &15.64 &1.59 &9.73 &3.95 &44.53 \\
LLaMA-3.2-90B-Vision-Instruction \cite{dubey2024llama} & Bbox & - & - & 16.21&1.75 &11.70 &4.53 & 48.29\\
InternVL-2-40B \cite{dubey2024llama} & Bbox & 1792 & 4096 &16.21 &1.79 &11.91 &4.63 &48.38 \\
\rowcolor[HTML]{FFF9E3}
GPT-4o \cite{achiam2023gpt} & Bbox & - & - &17.87 &3.21 & 12.87&6.49 & 49.85\\
\midrule
\rowcolor[HTML]{e9edf6}
\multicolumn{9}{c}{\textbf{\textit{Supervised Learning}}} \\
\midrule

Qwen2-VL-7B \cite{Qwen2VL} & Bbox & AnyRes & - &31.59 &9.11 &13.56 &90.32 &75.86 \\
LLaVA-1.6-13B \cite{liu2024visual} & Bbox & AnyRes & 576 &31.72 &9.35 &13.64 &90.71 &75.89 \\
VILA1.5-8B \cite{lin2023vila} & Bbox & 336 & 144 &31.87 &9.03 &13.79 &90.01 &75.95 \\
ViP-LLaVA-13B \cite{cai2024vip} & MContour & 336 & 576 &32.42 &9.97 &14.82 &91.44 &76.77\\
Alpha-CLIP-13B \cite{sun2024alpha} & Mask & 336 & 576 &35.68 &10.96 &16.11 &93.85 &77.66 \\
LLaVA-HR-X \cite{luo2024feast} & Bbox & 1024 & 1024 &35.97 &11.25 &16.57 &95.12 &78.08 \\
LLaMA-3.2-11B-Vision \cite{dubey2024llama} & Bbox & - & - &38.14 &12.87 &18.31 &99.11 &78.94 \\
\midrule
\rowcolor[HTML]{DAE8FC} 
\textbf{\NAME -8B (ours)} & Mask & 1024 & 1024 &\textbf{41.05} &\textbf{14.46} &\textbf{22.01} &\textbf{127.95} &\textbf{80.97} \\
\bottomrule
\end{tabular}%
}
\vspace{-1mm}
\caption{Comparison of the capabilities of \NAME and other related VLMs including both open-sourced models and \colorbox{api}{API-based models}. The column ``Visual Prompt'' indicates the format of input region-referral, ``Resolution'' indicates the encoding resolution for input images. We evaluate the models' performance under both the zero-shot and supervised learning settings.}
\label{tab:main}
\vspace{-4mm}
\end{table*}

%% file: sec/4_experiment.tex
\section{Experiments}
\label{sec:exp}
In this section, we present a comprehensive evaluation of our model, \NAME, in comparison with various state-of-the-art VLMs on the task of mask-referring, multi-grained image compositional captioning. 

\subsection{Baseline Models}
We selected a range of strong VLMs as baselines for both zero-shot and supervised learning settings to highlight \NAME's performance in mask-referring, multi-grained image compositional captioning. These include: Kosmos-2 \cite{peng2023kosmos}, Alpha-CLIP \cite{sun2024alpha}, Qwen2-VL \cite{Qwen2VL}, Ferret \cite{you2023ferret}, ViP-LLaVA \cite{cai2024vip}, LLaMA-3.2\cite{dubey2024llama}, GPT-4o \cite{achiam2023gpt}, ViP-LLaVA \cite{cai2024vip}, LLaVA-1.6 \cite{liu2024improved}, LLaVA-HR \cite{luo2024feast}, and the API-based model GPT-4o \cite{achiam2023gpt}.

\subsection{Implementation Details}
In our mixture-of-encoder architecture, we incorporate CLIP-ViT-L/14@336p \cite{radford2021learning} with an additional alpha channel -- similar to Alpha-CLIP \cite{sun2024alpha} -- for mask-aware low-resolution encoding. For high-resolution image encoding, we use ConvNeXt-XXL@1024p \cite{liu2022convnet} and SAM \cite{kirillov2023segment} encoders. As the language model decoder, we employ LLaMA-3.1-8B-Instruction \cite{touvron2023llama}. All encoder and decoder parameters are trainable, and we adopt the default hyperparameters from LLaVA-1.6 \cite{liu2024improved}. We trained our model on 8 Nvidia-A100 GPUs, and the whole training process took 38 hours.

\input{tables/fine-grained-results}
\subsection{Experimental Results on \DATA}
We evaluate models under both the zero-shot and supervised learning settings across multiple metrics. 
Table \ref{tab:main} presents the main results of \NAME alongside baseline models. From the zero-shot learning results, we observe that most models struggle to recognize both region referrals and attribute instructions precisely. Even strong open-source models like InternVL-2-40B and LLaMA-3.2-90B perform suboptimally on tasks introduced by \DATA, achieving only 11.7 and 11.9 in METEOR, and 5.5 and 4.63 in CIDEr, respectively. Additionally, the performance of the API-based model GPT-4 is still unsatisfactory in this context. In contrast, the supervised learning results demonstrate that models trained with our data exhibit significantly stronger performance. \NAME outperforms all other models due to the incorporation of mask-aware encoders and multiple high-resolution encoders, which enhance its ability to recognize region referrals and attribute-specific instructions effectively.

\subsection{Fine-Grained Evaluation}
In addition to traditional image captioning evaluation methods, we designed fine-grained evaluation techniques for the Attribute-Aware Regional Captioning (AARC) and Regional Dense Captioning (RDC) tasks. Following previous work \cite{you2023ferret,liu2024improved}, we use GPT-4 as a judge for the AARC task, where GPT-4 is fed the model's predictions, the image, and the ground truth to assess whether the model properly describes the attributes of the referred region. To better evaluate performance on the RDC task, we trained an LLaVA-1.6-13B model using the \DATA test set to predict bounding boxes for the region descriptions. We calculate Acc@0.5 to assess the quality of the predicted regional dense captions. From Table \ref{tab:eval}, we observe that \NAME achieves 56.84 on the AARC task and 83.49 on the RDC task, significantly outperforming the baseline models and indicating the superior capability of our model in these tasks. We provide the evaluation details in the appendix.

\subsection{Results on Referring Expression Generation Tasks}
To further evaluate our model's ability to recognize region referrals accurately, we conduct experiments on traditional REG tasks. Following the evaluation framework of previous works \cite{peng2023kosmos,yuan2024osprey}, we use the RefCOCOg test set and calculate METEOR and CIDEr scores for evaluation. Table \ref{tab:reg} presents the performance of different models compared to \NAME. Our model significantly outperforms previous models, indicating its superior capability in region description.

\input{tables/reg}
\input{tables/ablation}

%% file: tables/fine-grained-results.tex
\begin{table}[htbp]
\centering
\resizebox{\columnwidth}{!}{%
\begin{tabular}{l|c|c}
\toprule
\multirow{2}{*}{\textbf{Model}} & \multicolumn{1}{c|}{\textbf{AARC}} & \multicolumn{1}{c}{\textbf{RDC}} \\ \cline{2-3}
& GPT4-as-a-Judge $\uparrow$& Grounding Score (Acc@0.5) $\uparrow$\\ \midrule\midrule
ViP-LLaVA-13B \cite{cai2024vip} & 38.46 & 74.37 \\
Alpha-CLIP-13B \cite{sun2024alpha} &43.89 & 77.61 \\
LLaVA-HR-X \cite{luo2024feast} & 45.76 & 79.15 \\
LLaMA-3.2-11B-Vision \cite{dubey2024llama} & 50.26 & 81.01 \\
\midrule
\rowcolor[HTML]{ECF4FF}\textbf{\NAME-8B (ours)} & \textbf{56.84} & \textbf{83.49} \\
\bottomrule
\end{tabular}
}
\vspace{-1mm}
\caption{GPT-4-as-a-Judge score and Grounding score for Attribute-Aware Regional Captioning (AARC) and Regional Dense Captioning (RDC) tasks. The GPT-4-as-a-Judge score indicates the binary accuracy for correct attribute description, and the Grounding score is IoU Acc@0.5.}
\label{tab:eval}
\vspace{-4mm}
\end{table}

%% file: tables/reg.tex
\begin{table}[htbp]
\centering
\resizebox{\columnwidth}{!}{%
\begin{tabular}{l|c|cc}
\toprule
\multicolumn{1}{l|}{\multirow{2}{*}{\textbf{Model}}} & \multicolumn{1}{c|}{\multirow{2}{*}{\textbf{Region Referral}}} & \multicolumn{2}{c}{\textbf{RefCOCOg}} \\ \cline{3-4} 
\multicolumn{1}{c|}{} & \multicolumn{1}{c|}{} & METEOR & CIDEr \\ \midrule\midrule
GRIT \cite{muennighoff2024generative} & Bbox & 15.2 & 71.6 \\
Kosmos-2 \cite{peng2023kosmos} & Bbox & 14.1 & 62.3 \\
OMG-LLaVA \cite{zhang2024omg} & Mask & 15.3& - \\
GLaMM \cite{rasheed2024glamm} & Bbox & 16.2 & 105.0 \\
Osprey \cite{yuan2024osprey} & Mask & 16.6 & 108.3 \\ 
Alpha-CLIP+LLaVA \cite{sun2024alpha} & Mask & 16.7 & 109.2 \\ 
RegionGPT \cite{guo2024regiongpt} & Mask &16.9 &109.9\\
ControlCap \cite{zhao2025controlcap} & Bbox & 17.0 & 111.4 \\ \midrule
\rowcolor[HTML]{ECF4FF} \textbf{\NAME-8B (ours)} & \textbf{Mask} & \textbf{17.5} & \textbf{118.2} \\ \bottomrule
\end{tabular}%
}
\vspace{-1mm}
\caption{Region captioning performance evaluated on the test set of RefCOCOg.}
\label{tab:reg}
\vspace{-4mm}
\end{table}

%% file: tables/ablation.tex
\begin{table*}[t]
\centering
\resizebox{0.78\textwidth}{!}{
\begin{tabular}{l|ccccc} 
\toprule
\textbf{Method} & \textbf{ROUGE-L} & \textbf{BLEU-4} & \textbf{METEOR} & \textbf{CIDEr} & \textbf{BERT Score}  \\ 
\midrule 
\rowcolor[HTML]{ECF4FF}
\NAME & \textbf{41.05} & \textbf{14.46} & \textbf{22.01} & \textbf{127.95} & \textbf{80.97} \\ 
\midrule
\multicolumn{6}{c}{\textbf{\textit{Model Architecture}}} \\ 
\midrule
\NAME w/ LR Encoding Only & 37.92 & 11.67 & 17.86 & 97.62& 78.11  \\ 
\NAME w/ ConvNeXt & 39.87 & 13.42 & 20.97 & 109.25 & 79.83  \\ 
\NAME w/ SAM & 38.97 & 12.95 & 20.01 & 106.74 & 79.36  \\ 
\NAME + Self-Attn Fusion & 38.21& 12.26 & 19.73 & 104.85 & 79.24  \\ 
\NAME + Sequence Append Fusion& 36.11& 10.13 & 16.07 & 93.26 & 77.25  \\ 
\midrule
\multicolumn{6}{c}{\textbf{\textit{Region Referral}}} \\ 
\midrule
\NAME + Bbox & 37.10 & 11.12 & 17.21 & 96.85 & 77.91  \\ 
\NAME + MContour & 36.59 & 10.62 & 16.88 & 95.97 & 77.03  \\ 
\bottomrule
\end{tabular}}
\caption{Performance comparison of different model designs and region referrals.}
\label{tab:ablation}
\vspace{-2mm}
\end{table*}

%% file: sec/5_analysis.tex
\section{Analysis}
\label{sec:analysis}

\subsection{Ablation Study}
\noindent{\textbf{Models Design}.}
To better evaluate the effectiveness of our proposed model, we conduct an ablation study from the perspective of vision encoder fusion and different visual referring inputs. The ablation includes two parts, model design and region referral format. In model design, we investigate the effects of different visual encoder architectures and feature fusion strategies. From Table \ref{tab:ablation} we can observe that models with mask-aware low-resolution encoders only achieve lower performance compared to those that incorporate high-resolution encoders. Specifically, the \NAME with LR Encoding Only variant attains a CIDEr score of 97.62, which is significantly less than the full model's score of 127.95. This indicates that relying solely on low-resolution features is insufficient for capturing the detailed information necessary for high-quality caption generation.

When investigating different high-resolution encoder architectures, we find that \NAME w/ ConvNeXt Encoder outperforms \NAME w/ SAM Encoder, achieving a higher CIDEr score of 109.25 versus 106.74. This indicates that the combination of LR and HR encoder architecture significantly impacts model performance. In addition, the fusion strategy plays a crucial role: the \NAME + Self-Attention Fusion variant yields better results than the \NAME + Sequence Append Fusion. The latter underperforms with a CIDEr score of 93.26, suggesting that simply appending sequences is a less effective method for feature fusion.

\noindent{\textbf{The Impact of Referring Format}.}
In the region referral format ablation, replacing the precise segmentation masks with bounding boxes (\NAME + Bounding Box) and mask contours (\NAME + Mask Contour) leads to noticeable performance drops, with CIDEr scores of 96.85 and 95.97, respectively. This decline underscores the importance of using detailed mask information for accurately grounding visual content within the image.

Overall, the ablation study confirms that incorporating both low-resolution and high-resolution encoders, along with effective fusion strategies and precise region referrals, substantially enhances the model's ability to generate detailed and accurate captions. These results validate the effectiveness of our model.

\begin{figure*}[h!]
    \centering
    \includegraphics[width=0.84\textwidth]{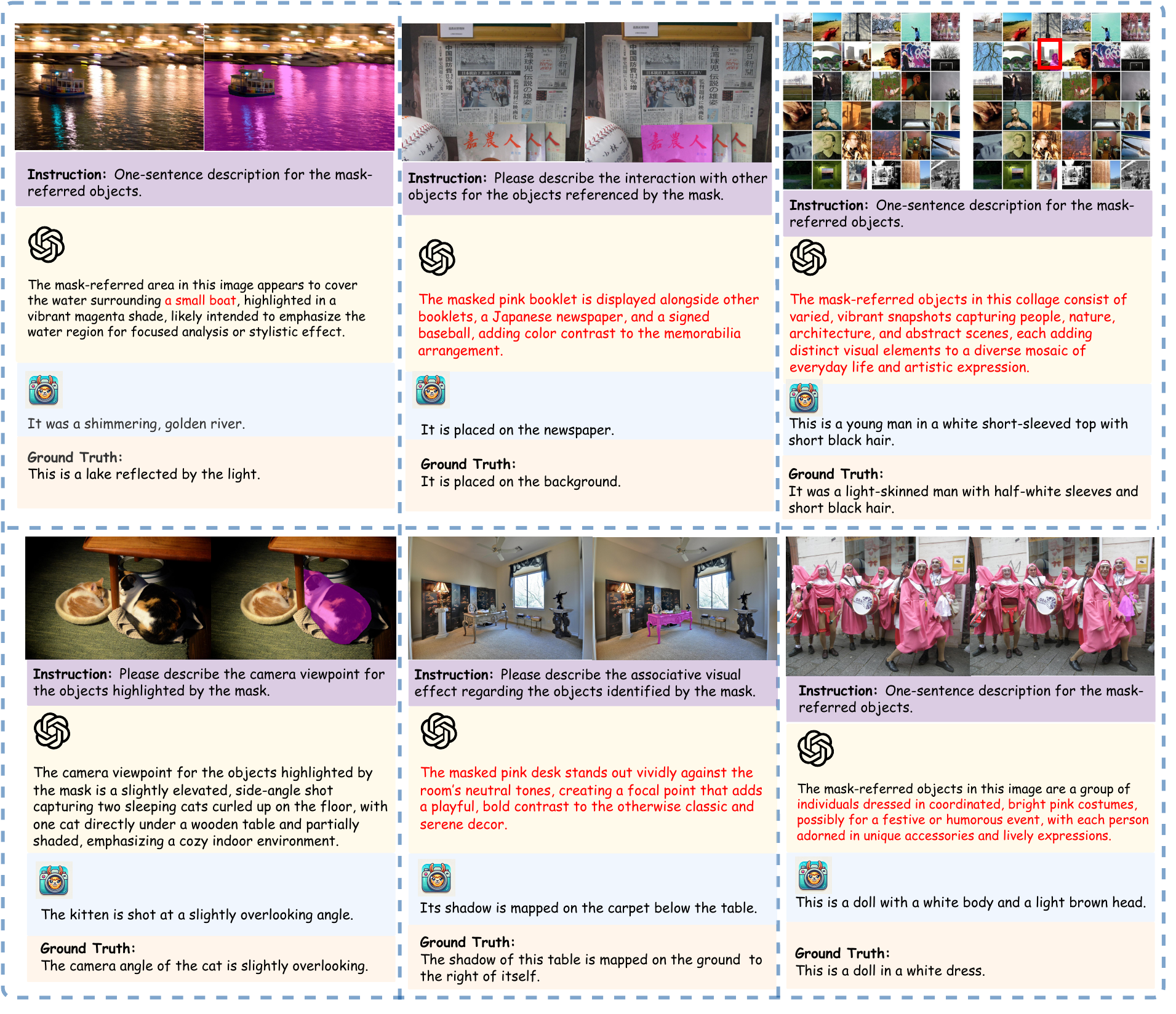}  
    \caption{\NAME provides accurate and concise descriptions focused on specified attributes and regions, while GPT-4o often misses fine-grained references and includes irrelevant information.}
    \label{fig:case}
    \vspace{-4mm}
\end{figure*}

\subsection{Qualitative Analysis}
Figure \ref{fig:case} compares the performance of different models. The examples illustrate that while GPT-4o can recognize some attribute instructions, it struggles with identifying fine-grained region references and often includes extraneous information unrelated to the specified attributes or regions. In contrast, our model, \NAME, excels in both attribute recognition and precise region localization, generating accurate and succinct descriptions focused exclusively on the specified regions and attributes, thereby reflecting a superior understanding of fine-grained compositional details.

%% file: sec/6_conclusion.tex
\section{Conclusion}
\label{sec:conclusion}
In this paper, we present \NAME, a vision-language model designed for mask-referring, multi-grained image compositional captioning. By combining a mask-aware encoder with high-resolution encoders, \NAME effectively perceives fine-grained details and accurately recognizes masked regions, outperforming models like GPT-4 and LLaMA-3.2 in detailed region-level tasks. To support our model, we created \DATA, a human-annotated dataset featuring 18 compositional aspects across diverse scenes and offering three levels of captioning granularity. This dataset fills a gap by emphasizing rich, attribute-aware descriptions of specified regions. Our work lays a foundation for advanced region-level understanding in vision-language models. We hope \NAME and \DATA become valuable resources for future research in detailed image perception and captioning.

%% file: sec/X_suppl.tex
\clearpage
\setcounter{page}{1}
\maketitlesupplementary

\section{More Quantitative Analysis of \DATA}
\label{sec:rationale}
In this section, we present additional quantitative analyses of \DATA to highlight the diversity and richness of compositional image captions.

\begin{figure}[htbp]
    \centering
    \includegraphics[width=0.95\linewidth]{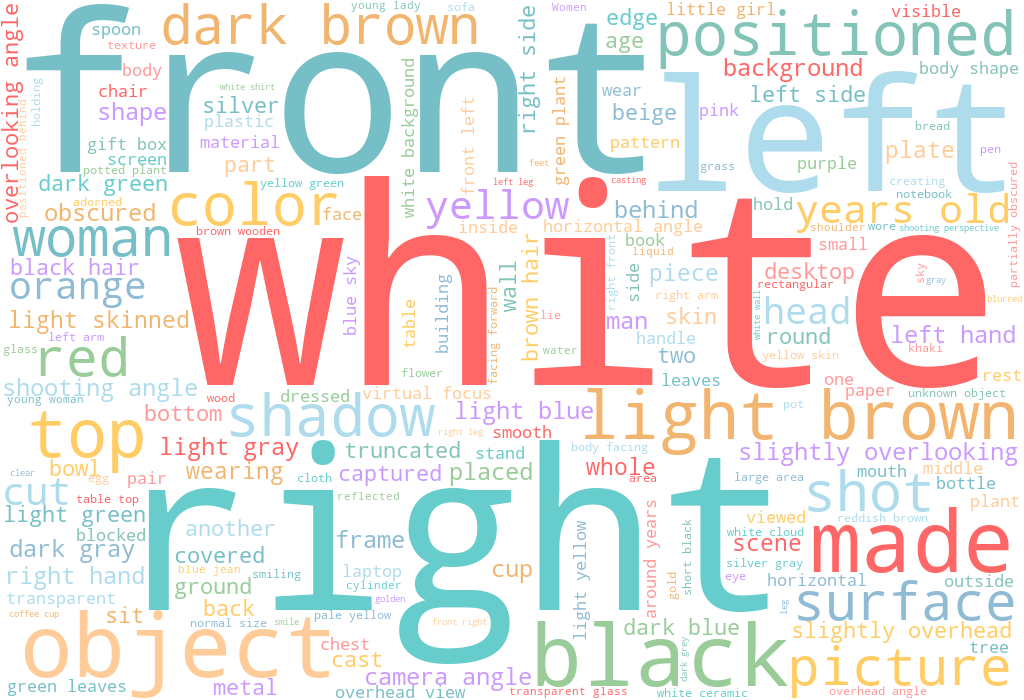}  
    \caption{Word cloud of key terms from the captions in \DATA, illustrating the diverse compositional region description of images. }
    \label{fig:count}
\end{figure}

Figure~\ref{fig:count} shows a word cloud generated from the captions in \DATA, illustrating the prominence of key descriptive terms across different regions in the dataset. Words like \textit{front}, \textit{white}, \textit{right}, and \textit{top} dominate the cloud, reflecting the dataset's emphasis on spatial positioning, attributes, and object relations. The diverse vocabulary extends to nuanced properties such as \textit{shadow}, \textit{material}, and \textit{texture}, which underline the dataset's compositional expressiveness. This variety ensures that models trained on \DATA are well-equipped to handle multi-faceted image regions with complex relationships.

\begin{figure}[htbp]
    \centering
    \includegraphics[width=0.95\linewidth]{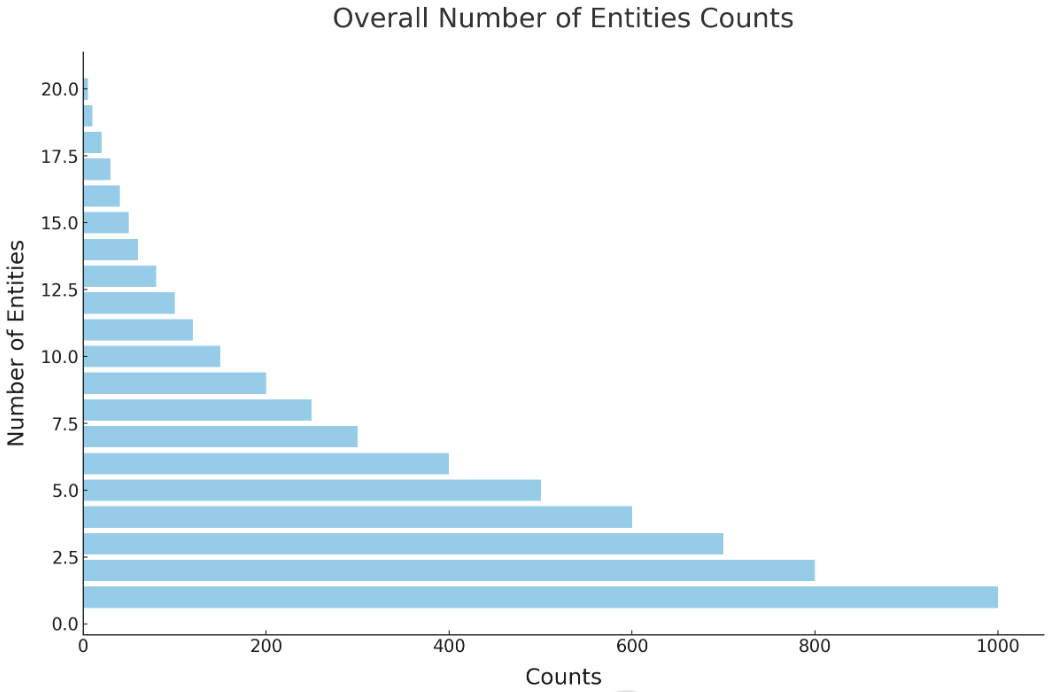}  

    \caption{Distribution of entities in \DATA.}
    \label{fig:count1}

\end{figure}

Figure~\ref{fig:count1} provides a histogram of attribute counts in \DATA, showcasing the distribution of entities across various descriptive aspects. The most frequent attributes include basic spatial and color properties, while less common features involve material and gesture-based descriptions. This balanced distribution supports comprehensive evaluation of model performance across common and rare descriptive cases.

\begin{figure}[htbp]
    \centering
    \includegraphics[width=0.95\linewidth]{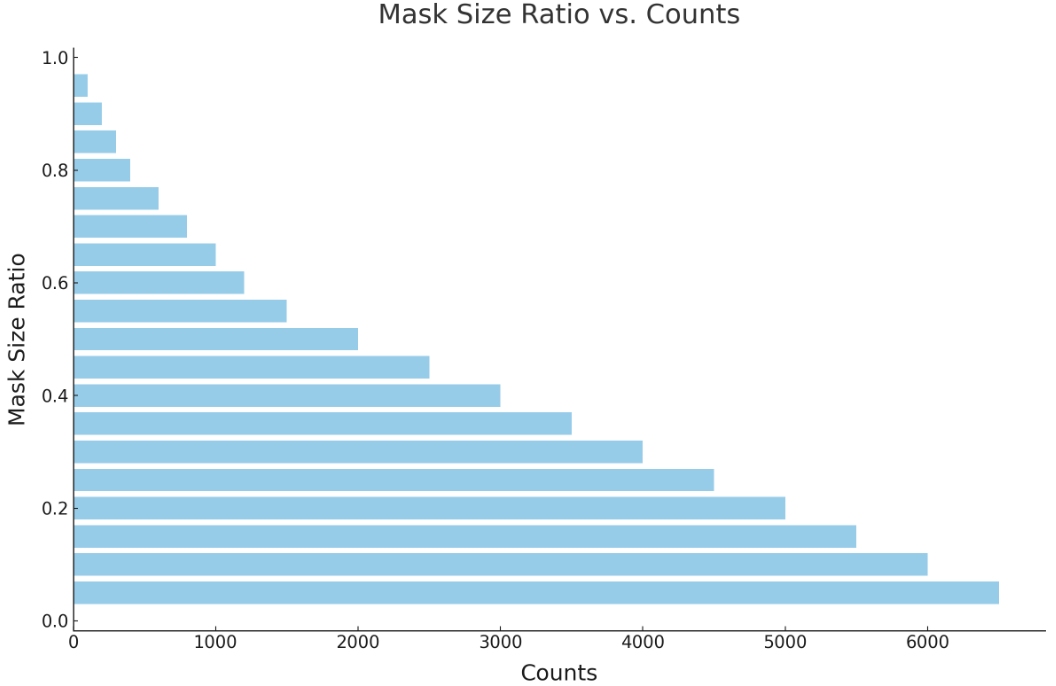}  

    \caption{Distribution of mask ratio in \DATA.}
    \label{fig:count2}

\end{figure}

Figure~\ref{fig:count2} depicts the distribution of mask size ratios in \DATA, representing the ratio of the region mask size relative to the entire image. This distribution highlights a diverse range of region sizes, from small, focused objects to large, encompassing areas. The gradual decline in counts as mask size ratios increase demonstrates the dataset's emphasis on evaluating both fine-grained and holistic compositional capabilities.

The quantitative insights presented here reaffirm the utility of \DATA as a robust benchmark for evaluating aspect-aware compositional reasoning in image captioning tasks.

\section{Explanation of Attributes}
\input{tables/aspects}
Table \ref{tab:aspect_explanations} provides detailed explanations of the attributes used in the Attribute-Aware Regional Captioning task.
\clearpage
\section{More Details and Cases}
\begin{figure*}[htbp]
    \centering
    \includegraphics[width=0.8\textwidth]{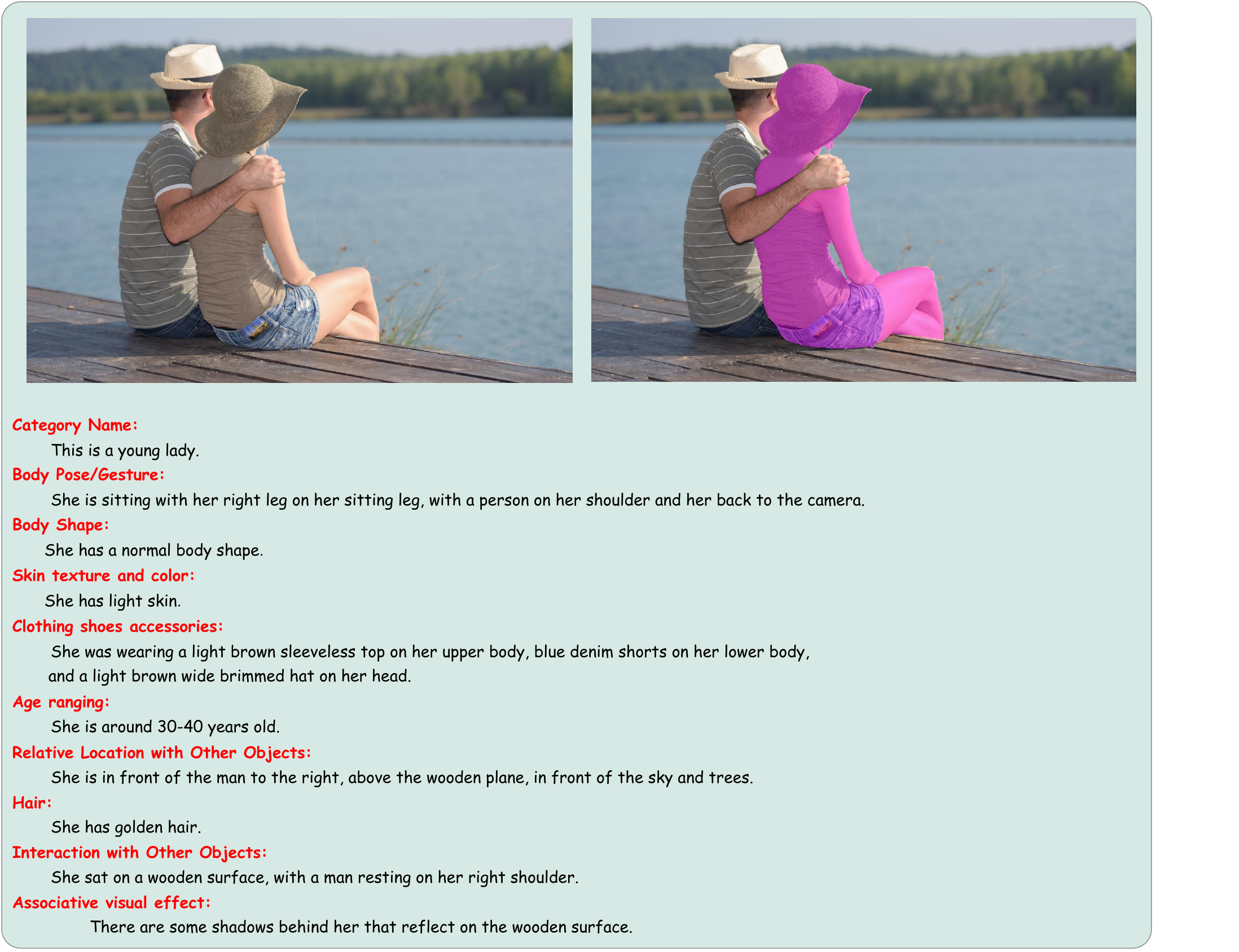}  
    \caption{Example for Attribute-Aware Regional Captioning task in \DATA.}
    \label{fig:case3}
\end{figure*}

\begin{figure*}[htbp]
    \centering
    \includegraphics[width=0.8\textwidth]{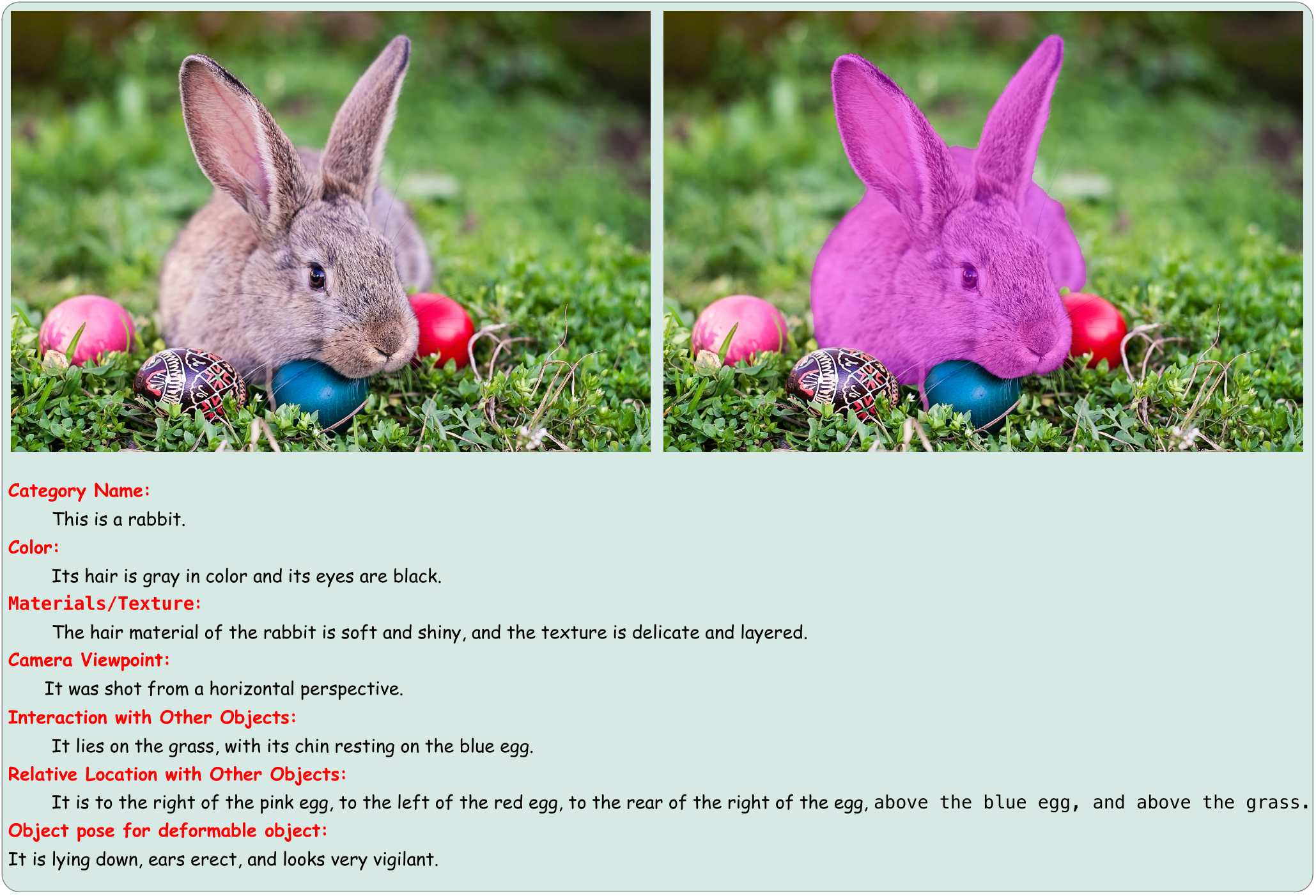}  
    \caption{Example for Attribute-Aware Regional Captioning task in \DATA.}
    \label{fig:case4}
\end{figure*}

\begin{figure*}[htbp]
    \centering
    \includegraphics[width=0.8\textwidth]{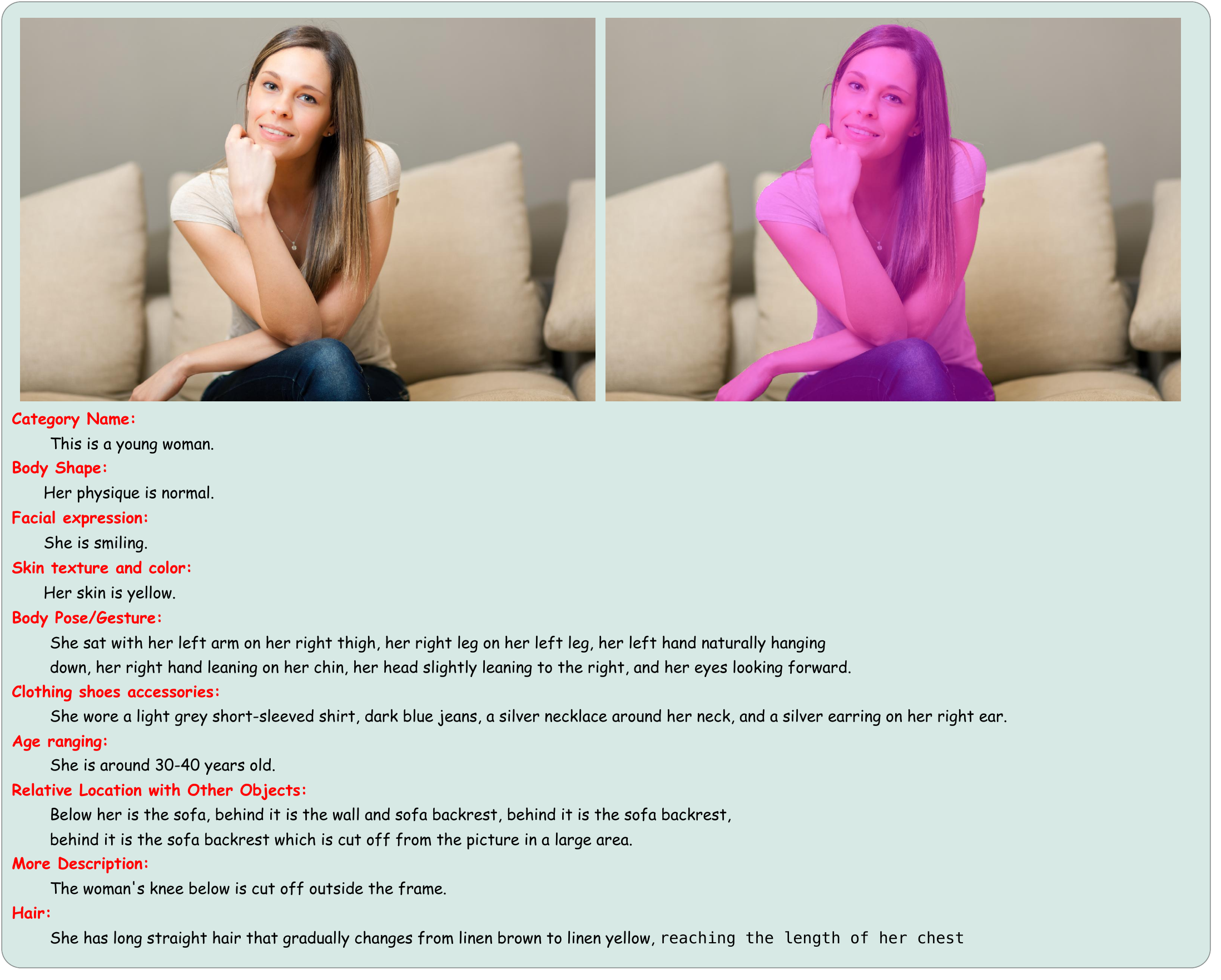}  
    \caption{Distribution of attributes in \DATA.}
    \label{fig:case1}
\end{figure*}

\begin{figure*}[htbp]
    \centering
    \includegraphics[width=0.6\textwidth]{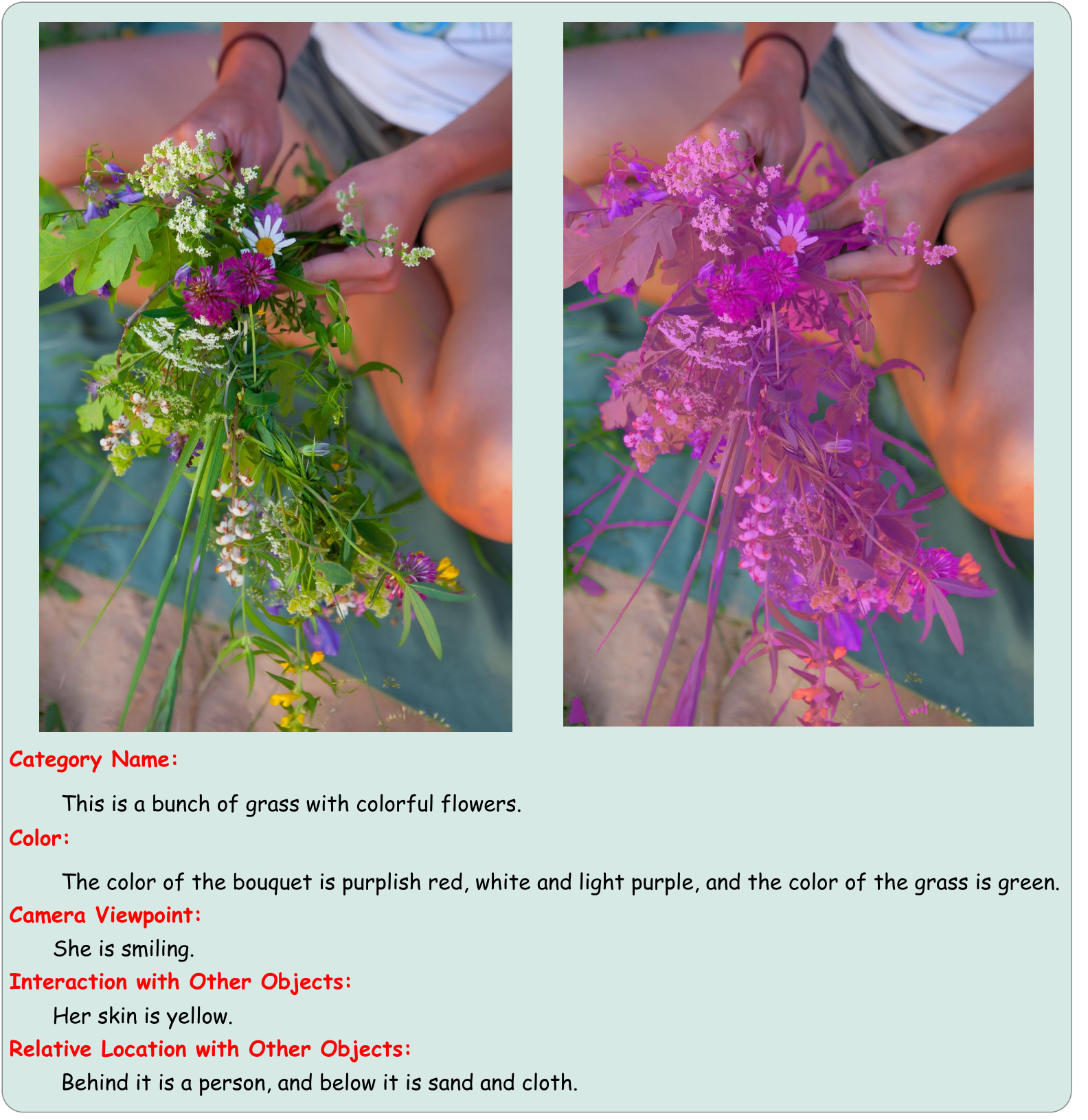}  
    \caption{Example for Attribute-Aware Regional Captioning task in \DATA.}
    \label{fig:case2}
\end{figure*}

\begin{figure*}[htbp]
    \centering
    \includegraphics[width=0.55\textwidth]{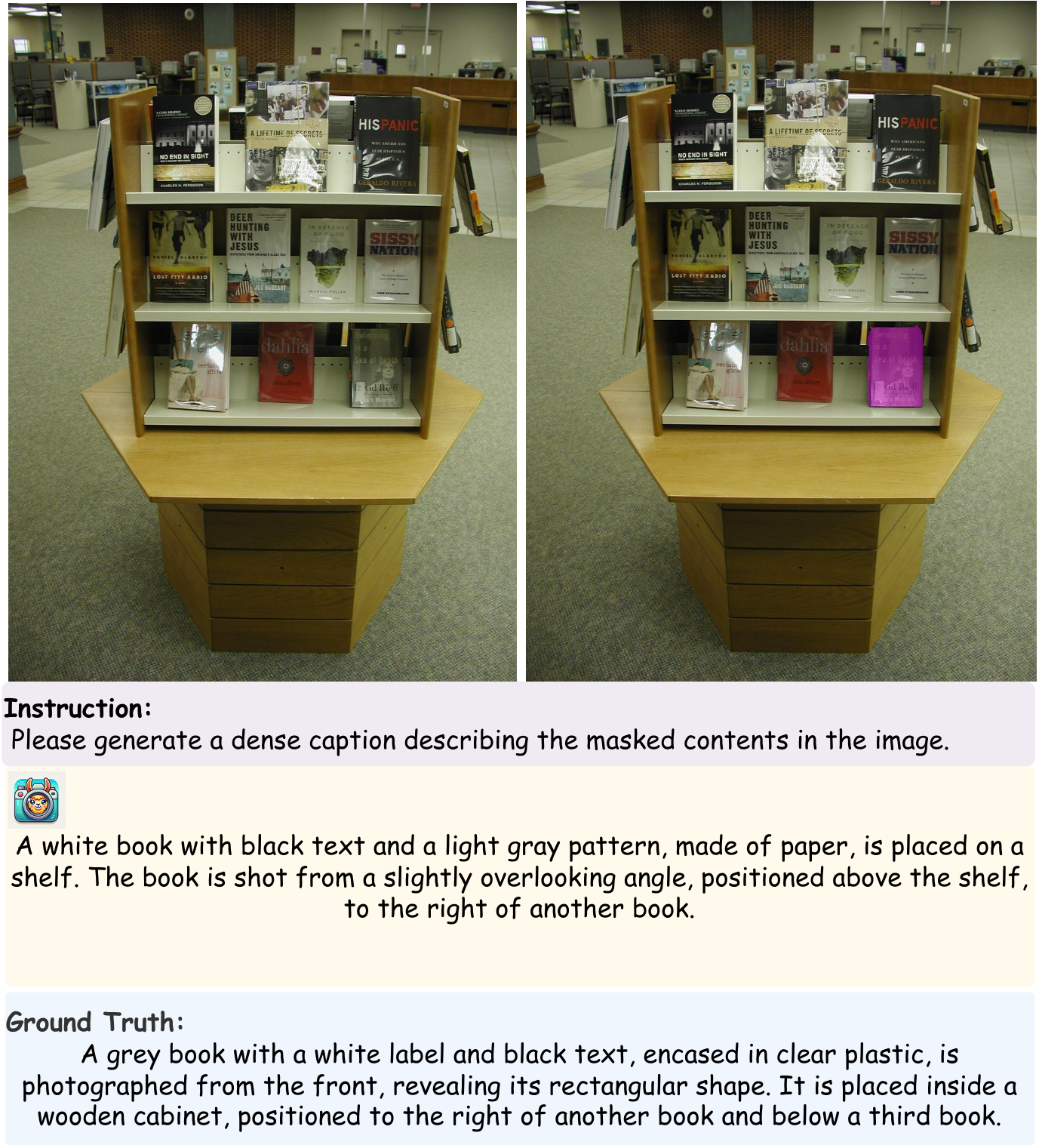}  
    \caption{Case study for Regional Dense Captioning task for \NAME.}
    \label{fig:case7}
\end{figure*}

\begin{figure*}[htbp]
    \centering
    \includegraphics[width=0.55\textwidth]{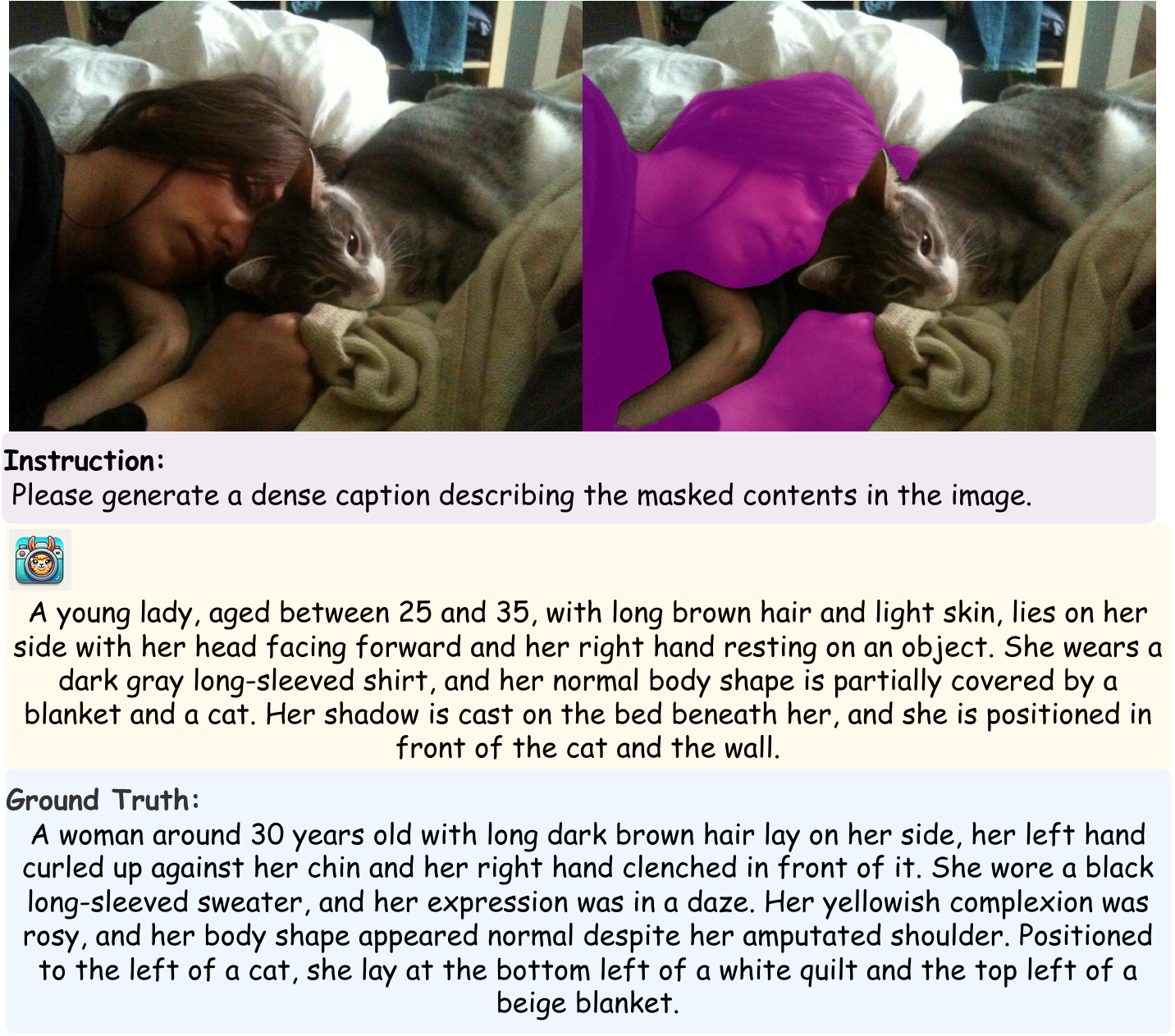}  
    \caption{Case study for Regional Dense Captioning task for \NAME.}
    \label{fig:case6}
\end{figure*}

\begin{figure*}[htbp]
    \centering
    \includegraphics[width=0.55\textwidth]{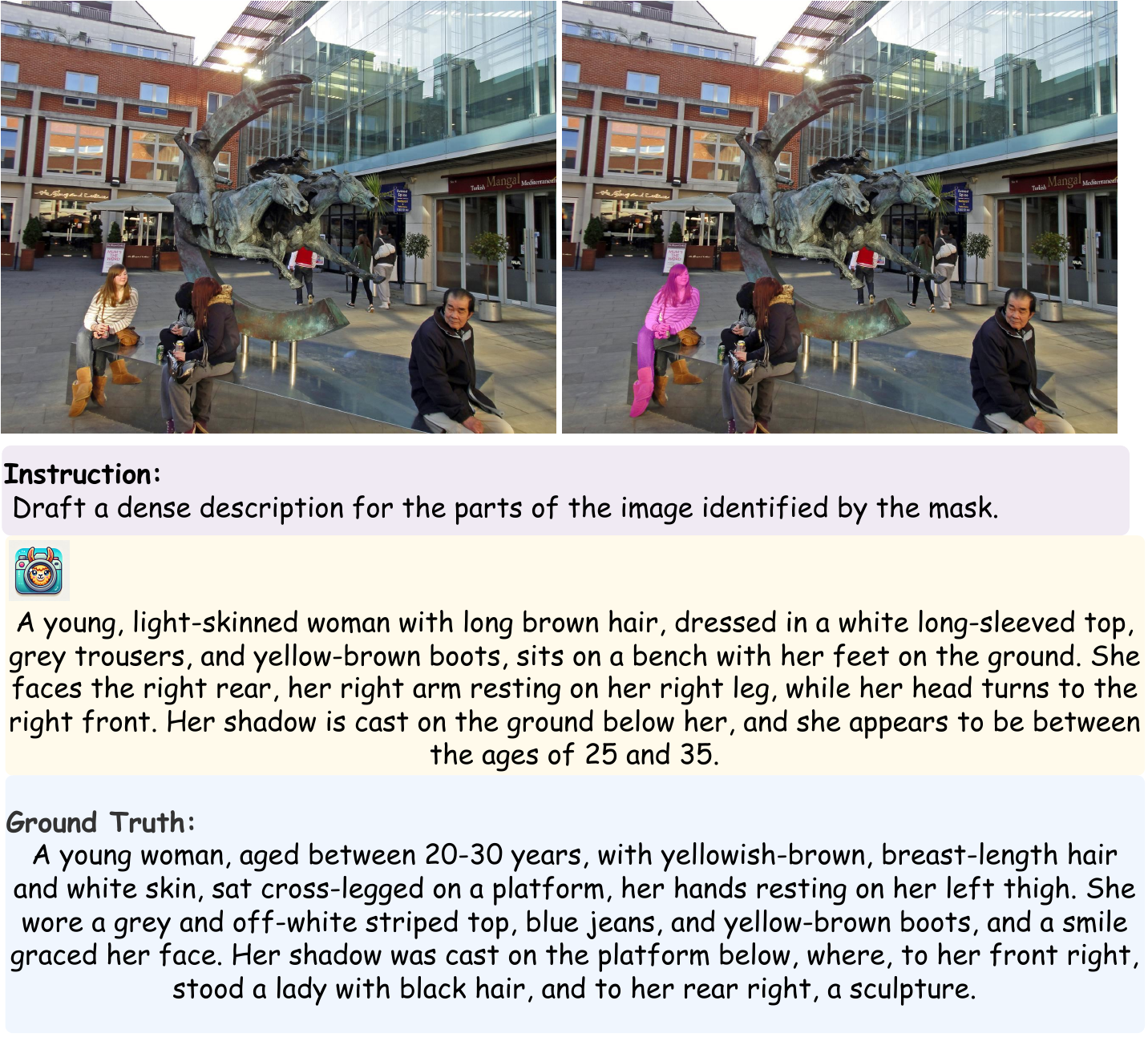}  
    \caption{Case study for Regional Dense Captioning task for \NAME.}
    \label{fig:case8}
\end{figure*}

\begin{figure*}[htbp]
    \centering
    \includegraphics[width=0.55\textwidth]{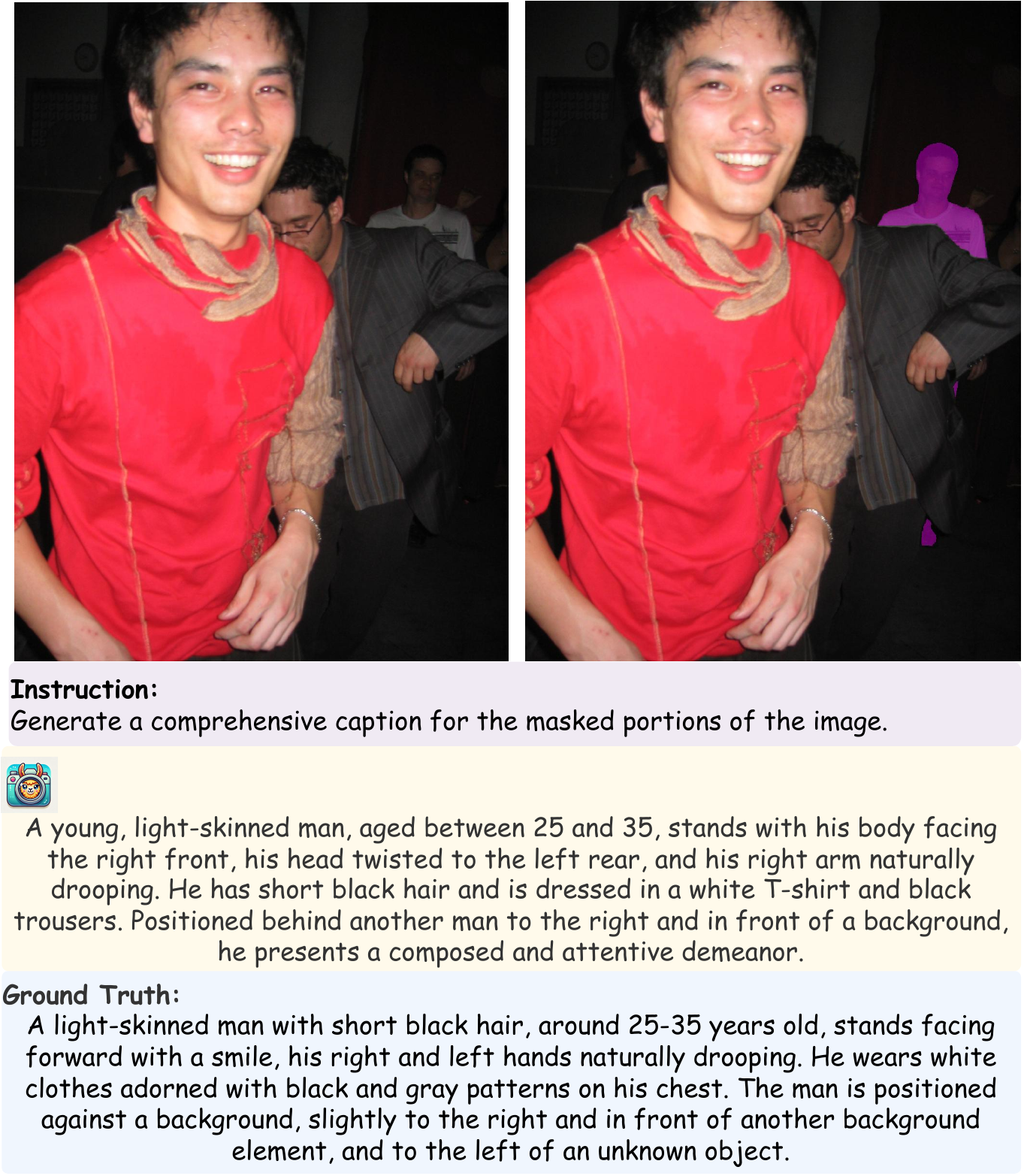}  
    \caption{Case study for Regional Dense Captioning task for \NAME.}
    \label{fig:case5}
\end{figure*}

In this section, we provide the prompt used for the GPT-4-as-a-Judge evaluation method. Additionally, we present more examples of our collected data in \DATA. Figures \ref{fig:case3} to \ref{fig:case2} illustrate the diversity and richness of our dataset. Figure \ref{fig:case7} to Figure \ref{fig:case5} showcase the predictions of \NAME for the Regional Dense Captioning task.
\begin{figure}[htbp] 
\centering
\begin{tcolorbox}[colback=white, colframe=SP, text width=0.85\columnwidth, title={\small Prompt for GPT4-as-a-Judge}, fontupper=\small, fontlower=\small]
Evaluator Instructions:\\
                You are an evaluator tasked with assessing the reasonableness of a model-generated caption for a specific attribute in a masked region of an image.

                You will be provided with:\\
                An image with a masked region (region of interest).\\
                A model-predicted caption.\\
                A reference description.\\

                Important Notes:\\
                The model's prediction does not need to exactly match the reference; it is acceptable as long as it reasonably describes the region and the attribute.\\
                The reference description serves as a suggestion or one possible answer, not an exact target.\\
                This is an open-ended generation task.\\
                Example: If the attribute relates to a person's age, and the prediction is "40-50 years old" while the reference is "45-50 years old," the prediction is considered reasonable.\\

                Your Task:\\
                Determine if the caption accurately and reasonably describes the expected attribute of the region of interest.\\
                Provide a binary answer ("Yes" or "No") based solely on whether the attribute description is reasonable.\\
                Please return "Yes" or "No" only, without any additional information.
                Please carefully examine all compositional details within the mask region!!\\
\end{tcolorbox}
\end{figure}



%% file: tables/aspects.tex
\begin{table*}[h!]
\centering
\resizebox{\textwidth}{!}{%
\begin{tabular}{>{\raggedright\arraybackslash}p{1cm}|>{\raggedright\arraybackslash}p{0.85\textwidth}}
\hline
\rowcolor[HTML]{EFEFEF} 
\textbf{No.} & \textbf{Aspect Explanation} \\
\hline
\hline
1  & \textbf{Category Name}: The general label or classification identifying the main subject in the image region, such as "dog," "tree," or "car." \\
\hline
2  & \textbf{Body Shape}: The form or outline of a living being's physique, including size, proportions, and overall build. \\
\hline
3  & \textbf{Skin Texture and Color}: The appearance of the skin's surface, detailing aspects like smoothness, roughness, and pigmentation. \\
\hline
4  & \textbf{Clothing, Shoes, Accessories}: The garments, footwear, and additional items worn or carried by a person, reflecting style or function. \\
\hline
5  & \textbf{Interaction with Other Objects}: How the subject is engaging with surrounding items, such as holding, sitting on, or leaning against something. \\
\hline
6  & \textbf{Body Pose/Gesture}: The positioning and movement of the subject's body parts, indicating action or posture. \\
\hline
7  & \textbf{Other Attributes}: Additional characteristics not covered by other aspects, like patterns, markings, or unique features. \\
\hline
8  & \textbf{Relative Location with Other Objects}: The spatial relationship between the subject and other elements in the scene, indicating proximity or arrangement. \\
\hline
9  & \textbf{Color}: The hues and shades present in the subject, contributing to its visual appearance. \\
\hline
10 & \textbf{Materials/Texture}: The substance an object is made of and the feel of its surface, such as metal, wood, smooth, or rough. \\
\hline
11 & \textbf{Camera Viewpoint}: The angle and perspective from which the image is captured, like frontal, side, aerial, or close-up views. \\
\hline
12 & \textbf{Associative Visual Effect}: Visual elements that create specific impressions or moods, such as shadows, reflections, or blurs. \\
\hline
13 & \textbf{Shape}: The external form or outline of an object, defining its geometry and structure. \\
\hline
14 & \textbf{Facial Expression}: The look on a person's face conveying emotion, like smiling, frowning, or surprised. \\
\hline
15 & \textbf{Hair}: The style, color, length, and texture of hair on a person or animal. \\
\hline
16 & \textbf{Age Ranging}: An estimation of the subject's age group, such as infant, child, teenager, adult, or elderly. \\
\hline
17 & \textbf{Object Pose for Deformable Object}: The positioning and form of objects that can change shape, like a twisted rope or crumpled paper. \\
\hline
18 & \textbf{Style}: The distinctive appearance or design of the subject, reflecting artistic trends, fashion, or aesthetic elements. \\
\hline
\end{tabular}%
}
\caption{Explanation of 18 Aspects in  Attribute-Aware Regional
Captioning task of \NAME.}
\label{tab:aspect_explanations}
\end{table*}